\documentclass[preprint,12pt,authoryear]{elsarticle}
\usepackage{amssymb}
\usepackage[utf8]{inputenc} 
\usepackage[T1]{fontenc}    
\usepackage{hyperref}       
\usepackage{url}            
\usepackage{booktabs}       
\usepackage{amsfonts}       
\usepackage{nicefrac}       
\usepackage{microtype}      
\usepackage{xcolor}         
\usepackage{amsmath}
\usepackage{graphicx}
\usepackage{makecell}
\usepackage{color}
\usepackage{float}
\usepackage{lineno}

\journal{Journal of Neural Networks}
\begin{document}

\begin{frontmatter}



\title{An Unsupervised STDP-based Spiking Neural Network \\Inspired By Biologically Plausible Learning \\Rules and Connections}

\tnotetext[mytitlenote]{Yiting Dong, Dongcheng Zhao  contributed equally to this work}
\author[1,3]{Yiting Dong}
\author[3]{Dongcheng Zhao}
\author[2,3]{Yang Li}

\author[1,2,3,4,5]{Yi Zeng}
\ead{yi.zeng@ia.ac.cn}

\address[1]{School of Future Technology, University of Chinese Acadamy of Sciences}
\address[2]{School of Artificial Intelligence, University of Chinese Academy of Sciences, Beijing}
\address[3]{Research Center for Brain-Inspired Intelligence, Institute of Automation, \\ Chinese Academy of Sciences (CAS), Beijing}
\address[4]{Center for Excellence in Brain Science and Intelligence Technology, \\ Chinese Academy of Sciences (CAS), Beijing}
\address[5]{National Laboratory of Pattern Recognition, Institute of Automation, \\ Chinese Academy of Sciences (CAS), Beijing}

\begin{abstract}
 The backpropagation algorithm has promoted the rapid development of deep learning, but it relies on a large amount of labeled data and still has a large gap with how humans learn. The human brain can quickly learn various conceptual knowledge in a self-organized and unsupervised manner, accomplished through coordinating various learning rules and structures in the human brain. Spike-timing-dependent plasticity (STDP) is a general learning rule in the brain, but spiking neural networks (SNNs) trained with STDP alone is inefficient and perform poorly. In this paper, taking inspiration from short-term synaptic plasticity, we design an adaptive synaptic filter and introduce the adaptive spiking threshold as the neuron plasticity to enrich the representation ability of SNNs. We also introduce an adaptive lateral inhibitory connection to adjust the spikes balance dynamically to help the network learn richer features. To speed up and stabilize the training of unsupervised spiking neural networks, we design a samples temporal batch STDP (STB-STDP), which updates weights based on multiple samples and moments. By integrating the above three adaptive mechanisms and STB-STDP, our model greatly accelerates the training of unsupervised spiking neural networks and improves the performance of unsupervised SNNs on complex tasks. Our model achieves the current state-of-the-art performance of unsupervised STDP-based SNNs in the MNIST and FashionMNIST datasets. Further, we tested on the more complex CIFAR10 dataset, and the results fully illustrate the superiority of our algorithm. Our model is also the first work to apply unsupervised STDP-based SNNs to CIFAR10. At the same time, in the small-sample learning scenario, it will far exceed the supervised ANN using the same structure.
\end{abstract}

%


\end{frontmatter}



\section{Introduction}
Simulating and designing a machine that thinks like a human is the ultimate goal of artificial intelligence. The vast majority of deep learning models rely on backpropagation algorithms, which require a large amount of labeled data to adjust parameters. However, obtaining labeled data is expensive. The backpropagation algorithm has a series of constraints, such as weight transport problem~\cite{lillicrap2016random}, and requires accurate gradient derivation, which is quite different from the learning process in the human brain. The human brain learns rapidly by relying on unsupervised local learning rules. Meanwhile, the traditional artificial neurons are far from the real spiking neurons which are rich in spatiotemporal dynamics~\cite{maass1997networks}. Spiking neurons receive input current and accumulate membrane potential, transmitting information through discrete spike sequences when the membrane potential exceeds the threshold. The spiking neural networks (SNNs) are more biologically plausible and energy efficient and have been widely used in various fields~\cite{fang2021brain,zhao2022brain,zhao2021spiking}.

\begin{figure}[t!]
	\centering
	\includegraphics[width=1.0\textwidth]{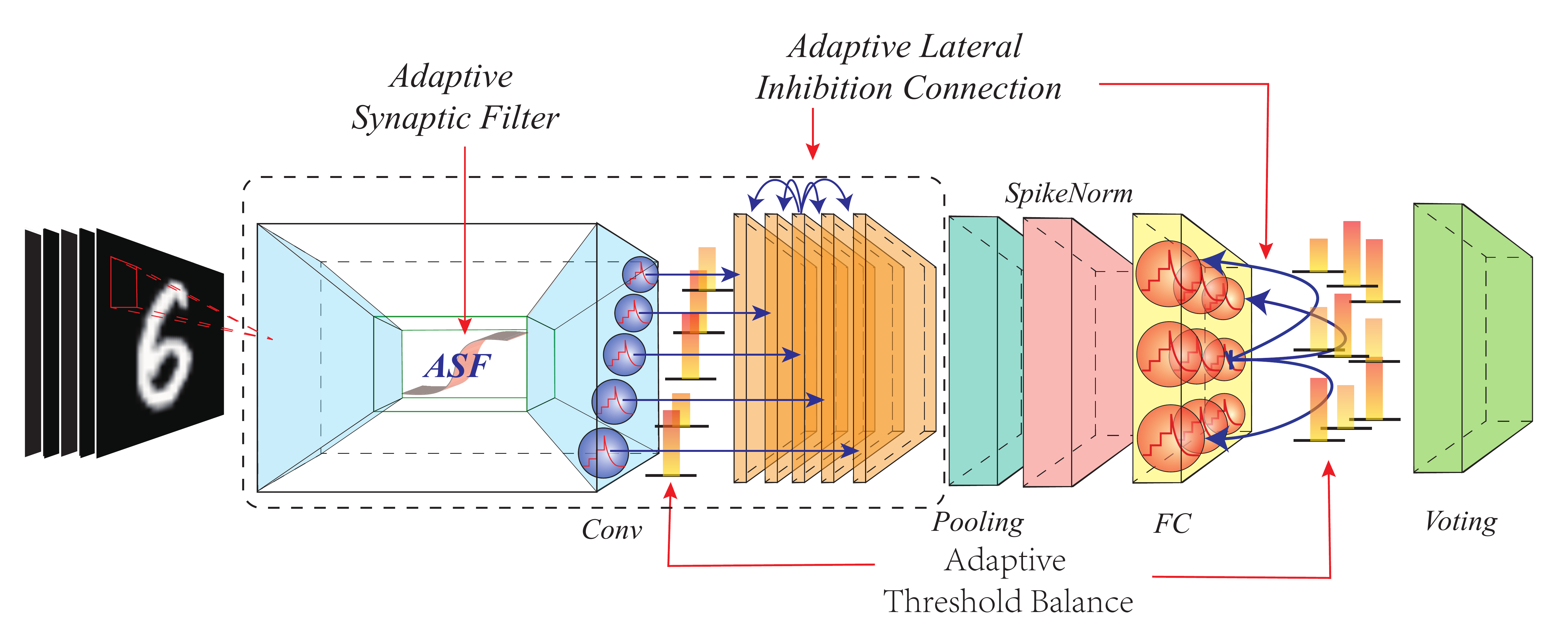}
	\caption{The backbone of our model, which introduces the adaptive synaptic filter, the adaptive threshold balance, and the adaptive lateral inhibitory connection to improve the information transmission and feature extraction of STDP-based SNNs. }
	\label{fig1}
\end{figure}

Training an efficient and robust spiking neural network is a critical problem many researchers have been paying attention to. Due to the non-differentiable characteristics of the spiking neural network, it is challenging to directly use the backpropagation algorithm for training, which significantly restricts the development of the SNNs. Many researchers take inspiration from the learning process in the human brain and design some biologically plausible learning rules to train SNNs. The synaptic plasticity of neurons is the neurological basis of learning and memory in the brain \cite{bi1998synaptic}. Spike Timing Dependent Plasticity (STDP) is a common learning rule that exists in multiple areas of the brain and plays a vital role in the brain’s perception and learning process. STDP influences the strength of synapses through the temporal relationship of pre- and postsynaptic spikes.

SNNs trained based on STDP still perform poorly due to the local optimization rule without global guided error compared with the backpropagation algorithm. This will lead to a lack of coordination and self-organization within and between layers of the model. Different parameter settings can easily lead to disordered spikes, making it challenging to transfer useful information. The human brain is not regulated by a single learning rule\cite{abbott2004synaptic}. The brain dynamically coordinates multiple learning rules and connections for rapid learning and inference. \textcolor{black}{In mammals, short-term synaptic plasticity (STP) is another essential learning rule.}  It lasts for a short time and adaptively controls the activity of different firing frequencies to regulate the information transmission better in a different layer\cite{ zucker2002short,citri2008synaptic, tauffer2021short}. Inspired by this, this paper designs an adaptive synaptic filter to help amplify the difference of the input current for better information transmission. Also, the adaptive spiking threshold is designed as the neuron plasticity to reduce the information loss during transmission. The adaptive lateral inhibition connection of different input samples is introduced to the spiking neurons of the same layer, which improves the self-organization ability of the model and enables the network to learn more abundant representations. Also, this paper extends the original STDP with sample temporal batched processing, which significantly accelerates the training process. To summarize, our key contributions are:
 \begin{enumerate}
 \item We propose the adaptive synaptic filter, adaptive lateral inhibitory connection, and the adaptive threshold balance to assist the training of unsupervised STDP-based SNNs, which significantly improves the representation ability of SNNs, alleviates the problem of repetitive features, and compensates for the input-output mismatch between layers.

 \item We extend the original STDP by integrating multiple samples and different moments into a batch (STB-STDP), which significantly speeds up and stabilizes the training process. 

 \item Experimental results on MNIST and FashionMNIST show that our algorithm achieves the state-of-the-art performance of unsupervised spiking neural networks. At the same time, the performance of SNNs in complex scenarios is improved, allowing unsupervised SNNs to show excellent performance in CIFAR10 and small-sample training scenarios.
\end{enumerate}

\section{Related Work}
\textcolor{black}{ The training of spiking neural networks is currently divided into three categories, conversion-based, backpropagation-based, and brain-inspired algorithm learning rules based.}

 By exploring the relationship between the spike activation and the artificial activation function, the real value of the artificial neural networks (ANNs) can be approximated to the average firing rates of SNNs. As a result, an alternative way is explored to add constraints on the weights of well-trained ANNs to convert them to SNNs \cite{diehl2015fast,li2021bsnn,han2020deep}. Although these converted methods make SNNs show excellent performance in more complex network structures and tasks, it does not fundamentally solve the training problems of SNNs. Other researchers introduce the surrogate gradient to make the backpropagation algorithm can be directed used in the training of SNN~\cite{lee2016training,wu2018spatio,wu2019direct,shen2021backpropagation}. However, as said before, the backpropagation algorithm is implausible and far from how the brain learns.

Since STDP is a ubiquitous learning rule in the brain, many researchers trained spiking neural networks based on STDP. \cite{querlioz2013immunity} tried a two-layer fully connected SNN using a simplified unsupervised approach of STDP. \cite{diehl2015unsupervised} used an unsupervised STDP method with two layers of activation and inhibition. Notably, despite two layers, only one has trainable parameters.  \cite{kheradpisheh2018stdp} used hand-designed DoG filters for feature extraction, STDP to train convolutional layers, and SVM as the classifier. The convolution kernels of each layer are designed individually. Only the training in the intermediate convolutional layers is unsupervised. However, due to the local optimization property of STDP, it tends to perform poorly on deep networks, so many researchers have tried to introduce supervisory signals to guide STDP tuning based on the global feedback connections~\cite{zhao2020glsnn}, equilibrium propagation~\cite{zhang2018plasticity}, backpropagation~\cite{liu2021sstdp}, and the dopamine-modulated~\cite{hao2020biologically}. Some methods combine STDP with backpropagation for hybrid training. Such as \cite{liu2021sstdp} and \cite{lee2018training}, they both first performed STDP training to extract weights with better generalization. The training of supervised backpropagation is then performed to obtain better performance. 
 
 
Lateral inhibition is usually used to help neurons achieve mutual competition mechanism \cite{heitzler1991choice,amari1977dynamics,blakemore1970lateral}. The lateral inhibition mechanism has been tried to be added to the training of spiking neural networks. \cite{diehl2015unsupervised} tried to use a static lateral inhibition mechanism, so that the firing neurons can inhibit other non-spiking neurons by reducing the membrane potential. \cite{cheng2020lisnn} help the spiking neural network to have stronger noise-robustness by introducing lateral inhibitory connections. 
\section{Backgrounds}
\subsection{Neuron Model}

\textcolor{black}{The leaky integral-and-fire (LIF) neurons \cite{dayan2005theoretical} are the most commonly used computational model in the SNNs.} LIF neurons receive the pre-synaptic spikes as the input currents and accumulate them on the decayed membrane potential. When the membrane potential reaches the threshold, the neuron releases a spike with the membrane potential reset to the resting potential $u_{reset}$. Here we set $u_{reset}=0$. The details are shown in Equation~\ref{eq1}:
\textcolor{black}{
\begin{equation}
	\label{eq1}
	\begin{aligned}
		&s=0   \quad \tau \frac{d u }{d t}=-u+Ri  ,\quad &if\ 		u <u_{thresh} \\
		&s=1 \quad u=0,\quad &if \ u\geq u_{thresh}
	\end{aligned}
\end{equation}
}

where $u$ is the membrane potential. $u_{thresh}$ is the threshold for this neuron. $\tau$ is the time constant.   $i $ is the input current. We denote $i =\sum\limits_{j}w_{ij}s_{j}$. $s  $ is the spikes from pre-synaptic neuron. $w _{ij}$ is the strength of synapses. $R$ is resistance. 

In order to facilitate the calculation, we convert the differential formula into a discrete representation as shown in Equation~\ref{eq3}, where $C$ is capacitance, which we set equal to 1.
 
\begin{equation}
	\begin{aligned}
		&s^{(t)}=0 \quad u^{(t)}=(1-\frac{1}{\tau})u^{(t-1)}+\frac{1}{C}i ^{(t)} ,\quad &if\ 		u^{(t)} < u_{\mathit{thresh}} ^{(t)}\\
		&s^{(t)}=1 \quad u^{(t)}=0,\quad &if \ u^{(t)}\geq u_{\mathit{thresh}}^{(t)}
	\end{aligned}
	\label{eq3}
\end{equation}
 
where $u^{(t)}$ is the membrane potential at the time $t$. $u_{thresh}^{(t)}$ is the threshold for this neuron at the time $t$.   $i ^{(t)}$ is the input current at the time $t$. 
\subsection{STDP Algorithm}
In this paper, we improved the commonly used unilateral STDP.  For conventional  STDP, as seen in Equation~\ref{eq4}, the modification of weights is determined by the time interval of the pre- and post-synaptic spikes. The larger the time gap, the less correlated the two spikes are and the less affected the synaptic weights.
 
\begin{equation}
	\begin{split}
		&\Delta w_{j}=\sum\limits_{f=1}^{N} \sum\limits_{n=1}^{N} W(t^{f}_{i}-t^{n}_{j})\\
		&W(\Delta t)= 
		\begin{split}
			A^{+}e^{\frac{-\Delta t}{\tau_{+}}}-x_{\mathit{offset}}\quad if\; \Delta t>0&\\ 
		\end{split}
	\end{split}
	\label{eq4}
\end{equation}
 
where $\Delta w_{j}$ is the modification of the synapse $j$, $W(\Delta t)$ is the STDP function. For unilateral STDP, focus only on the presynaptic spikes before the firing of the postsynaptic spikes, which makes the synapse strength continue to grow \cite{lee2018training}. So $\textcolor{black}{x_{\mathit{offset}}}$ is to determine whether the modification are potentiated or depressed.

\textcolor{black}{For the efficient implementation, we take another form of STDP using eligibility traces \cite{lee2018training,izhikevich2007solving,zenke2015diverse}. As shown in the Equation \ref{eq42} , $x_{trace}^{(t)}$ accumulates presynaptic spikes and gradually decays over time.}

\begin{equation}
	\begin{split}
		&\Delta w_{j}=\textcolor{black}{x_{\mathit{trace}}^{(t)}}-\textcolor{black}{x_{\mathit{offset}}}\quad if \:\:\textcolor{black}{ x_{\mathit{postspike}}=1}\\
		&\begin{split}
			\textcolor{black}{x_{\mathit{trace}}}^{(t)}=\lambda_{+}\textcolor{black}{x_{\mathit{trace}}^{(t-1)}}+\textcolor{black}{x_{\mathit{prespike}}}
		\end{split}
	\end{split}
	\label{eq42}
\end{equation}
When the neuron fires at time $t$,\textcolor{black}{ $x_{\mathit{postspike}}=1$ }. When recieves a spike at time $t$, \textcolor{black}{$x_{\mathit{prespike}}=1$}. We denote $\lambda_{+}=1-\frac{1}{\tau+}$.

In experiments, we improve STDP and propose a sample-temporal batch STDP (STB-STDP) algorithm. More details are in section \ref{stbstdp}.

\section{Proposed Algorithms}
\begin{figure}[h]
	\centering
	\includegraphics[width=0.8\textwidth]{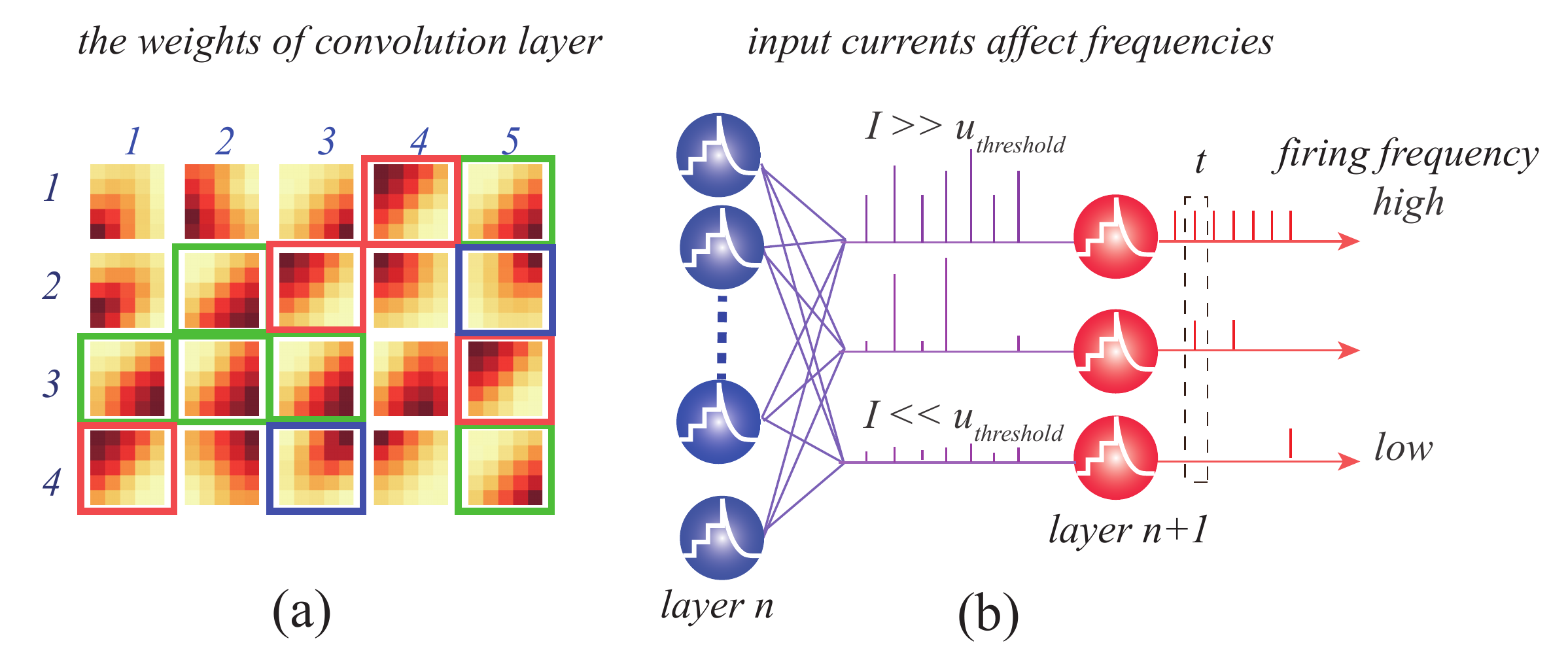}
	\caption{\textcolor{black}{(a) Features of the convolutional layers trained with only STDP, the same features are marked
		with the same color. (b) SNNs trained only with STDP cannot properly regulate the information transmitted by the spikes, which will lead to sparse or frequent spikes.}}
	\label{weightfrequency}
\end{figure}

To be more explicit about the problem we focus on, we built a one-layer convolutional spiking neural network and used the STDP algorithm, described in Equation \ref{eq42} , to train in MNIST dataset until convergence. This convolutional layer consists of a 5x5 convolution kernel with 20 channels. As shown in Figure~\ref{weightfrequency}(a), we show the synaptic weights of the convolutional layers after training. It can be observed that a large number of repeated convolution kernels appear. The same convolution kernel features are boxed in the same color. Repeated convolution kernels will affect the effective feature representation \cite{glorot2010understanding}.

On the other hand, neurons from different layers may also work in a disordered way.  
Since the algorithm lacks a global guided signal, the neuron can not judge whether it is a suitable firing rate. Therefore This will cause the adjacent next layer of neurons to receive too high or too low input current, which makes the last layer's neurons fire unstably. As shown in Figure \ref{weightfrequency}(b), The membrane potential will take longer to build up if the input current is too small, leading to delays in the transmission of information. When the firing frequency is significant, the neuron fires nearly all the time, which will damage the effective information representation.
Different parameter settings can easily lead to disordered spikes, making it challenging to transfer useful information.
\label{meth}

\textcolor{black}{To alleviate the above problems, we propose three adaptive algorithms}. \textcolor{black}{To address the problem of repetition of neuron features within a layer, we propose an Adaptive Lateral Inhibitory Connection (ALIC), different from static lateral inhibition \cite{diehl2015unsupervised}, which provides a way to coordinate neurons by automatically selecting those that need to be inhibited. }Next, we use Adaptive Threshold Balance (ATB) to solve the mismatching of input and output ranges between adjacent layers. In order to make the spikes firing more stable, we designed Adaptive Synaptic Filter (ASF) inspired by STP. Finally, we propose STB-STDP, which combines spatial and temporal information into a single batch.
\subsection{Adaptive Lateral Inhibitory Connection}
We introduce lateral inhibition to solve the problem of neurons in the same layer tending to have the same weights. Interaction between neurons is enabled by lateral inhibition, which allows firing neurons to dominate this input by inhibiting the firing of other neurons. The dominant neuron is more likely to experience learning according to STDP, and sensitivity to its input gradually increases. Conversely, non-dominant neurons will not be sensitive to this input. This prevents neurons from convergent towards the same weight.
\textcolor{black}{static lateral inhibition, however, is usually a constant set manually. }
\begin{equation}
	inh_{ij}^{(t)}=\alpha
\end{equation}
where $inh_{ij}^{(t)}$ denotes the inhibition received by the neuron $i,j$.  $\alpha$ is a parameter defining the degree of inhibition. The same inhibiting degree is maintained across all neurons and inputs. As a part of the network, the lateral inhibitory connections are static, with the same weights assigned to each neuron. However, this is not reasonable. \textcolor{black}{Evidence from neuroscience suggests that for better inhibition, it does not exert inhibition on all neurons, but on those with relevant activity \cite{linster2005computational,kuffler1953discharge,arevian2008activity}.} Thus, we expect a dynamic structure of lateral inhibitory connections to produce various structures for different inputs. Such a dynamic structure would help enhance coordination between neurons.

To this end, we introduce Adaptive Lateral Inhibitory Connection (ALIC), as shown in Figure \ref{asfatbalic} . \textcolor{black}{We designed a dynamic structure of lateral inhibition. we determine which neurons may fire by setting a threshold and inhibit the membrane potential for these neurons.  } As shown in the Equation \ref{inh}, we choose the maximum input current as the reference. Inhibition depends on maximum current for different inputs. Threshold is set with $i_{thresh}=\frac{i^{(t)}_{max}}{2}$

\begin{equation}
	\begin{split}
		inh^{(t)}_{ij}&=\alpha_{inh}\ (\max\limits_{b,c,w,h}i^{(t)} )(1-\hat{s}_{ij}^{(t)} )\\
		&=\alpha_{inh}\ (\max\limits_{b,c,w,h}\sum\limits_{i,j}w^{(t)}_{ij}\hat{s}_{ij}^{(t)})(1-\hat{s}_{ij}^{(t)} )\quad if \quad i^{(t)}_{ij}>i_{\mathit{thresh}} \\
	\end{split}
	\label{inh}
\end{equation}
\textcolor{black}{where $inh^{(t)}_{ij}$ is the inhibition for the neuron at the position of $i,j$ at the time $t$. $\alpha_{inh}$ is a coefficient that adjusts the degree of inhibition. $i^{(t)}$ denotes the input current, which obtained by synaptic weight $w_{ij}$ and spikes $\hat{s}_{ij}^{(t)}$. The maximum is selected from all the neurons in batch, where $b$ denotes batch, $c,w,h$ denotes the channel and size of output. $(1-s_{ij}^{(t)})$ allows the inhibition to act only on neurons that are not firing at this moment. } We adopt the winner-take-all strategy, randomly take out a firing neuron, and set the remaining spikes to 0, $ \hat{s}_{ij}^{(t)}=wintakeall(s_{ij}^{(t)})$.
A more detailed analysis of section \ref{analysis} shows that ALIC actually improves the performance of the network.

\subsection{Adaptive Threshold Balance}
By introducing an adaptive threshold method, we are able to eliminate the mismatch between input and output between layers. \textcolor{black}{The input current varies from different inputs, leading to spikes firing variability. }
The current may be too small to reach the threshold, delaying the transmission of information and increasing the network delay.
A large value may also be well above the threshold, increasing firing frequency. The portion of current above the threshold will be lost due to the reset of the membrane potential. Therefore, a dynamic threshold method is needed so that the threshold can be adaptively changed according to the magnitude of the current. Due to neurons' inherent plasticity, the adaptive threshold can reduce the loss during transmission and facilitate the expression of more precise information\cite{wilent2005stimulus, huang2016adaptive}. We introduce an adaptive threshold balancing (ATB) mechanism.

The convolutional and fully connected layers play different roles in the network. The convolutional layer extracts features, while the fully connected layer exhibits feature selectivity sensitive to different features. Therefore, we employ variable threshold methods at layers.
For convolutional layers, ATB set threshold positively related to the maximum input current, as shown in Equation \ref{eq_th}.  
where $u_{thresh}^{(t)}$ denotes the threshold of neuron at time $t$.
$\beta_{thresh}$ is a parameter controlling the threshold scale. The maximum current is selected from the maximum input value of all neurons in the convolutional layer. $c,w,h$ represents the number of channels in the convolutional kernel and the size of the output, respectively.  The input current of each neuron is obtained from the synaptic weights and the input spikes.
ATB ensures that no information is lost due to excessive current. Meanwhile, it allows spikes to be transmitted for a limited time.

\begin{equation}
	\label{eq_th}
	\begin{split}
		u^{(t)}_{\mathit{thresh}}=\beta_{\mathit{thresh}}\ (\max\limits_{c,w,h}i^{(t)} )
		=\beta_{\mathit{thresh}}\ (\max\limits_{c,w,h}\sum_{i,j} w^{(t)}_{ij}\hat{s}_{ij}^{(t)})
	\end{split}
\end{equation}

For fully connected layers, ATB improves the method in \cite{diehl2015unsupervised}. \textcolor{black}{Where $u^{t,j}_{\mathit{thresh}}$ denotes the threshold of the neuron $j$ at time $t$. When a neuron fires a spike, we increase the threshold $\theta_{plus}^{t,j}$, making the next firing more difficult. When $\theta_{plus}^{t,j} $ reaches $\gamma$, all thresholds reduce the difference between maximum threshold and $\gamma$. As shown in Equation ~\ref{eq_th2} . Where $\theta_{\mathit{init}}$ is the initial value. $\theta_{\mathit{plus}}$ is the increment of $u^{t,j}_{\mathit{thresh}}$. $\alpha_{plus}$  is a coefficient controlling the growth rate.}

\begin{equation}
	\label{eq_th2}
	\begin{split}
		&u^{t,j}_{\mathit{thresh}}=\theta_{\mathit{init}}+ \theta_{\mathit{plus}}^{t,j}\\
		&\theta_{\mathit{plus}}^{t,j}=\theta_{\mathit{plus}}^{t-1,j}+\alpha_{\mathit{plus}}\sum\limits_{b,t} \hat{s}^{(t)}_{j}-\theta_{\mathit{bias}}^{t,j}\\
		&\theta_{\mathit{bias}}^{t,j}=\left\{
		\begin{aligned}
			&  \max\limits _{j} u_{\mathit{thresh}}^{t-1,j}-\gamma&\quad if \quad \max\limits_{j} u_{\mathit{thresh}}^{(t-1)}>\gamma \\		
			& 0						   &\quad if \quad \max\limits_{j} u_{\mathit{thresh}}^{(t-1)}<\gamma		
		\end{aligned} \right.
	\end{split}
\end{equation}

In this manner, a dynamic balance is established, where each neuron is measured for relative sensitivity. When $\gamma$ is reached, the whole shifts downward. All thresholds are limited within a range. Thus, the difference between the thresholds is indicative of the neuron's sensitivity. \textcolor{black}{Neurons with higher thresholds have lower sensitivity.} This dynamic balance prevents a single neuron from dominating and ensures that each neuron has an opportunity to fire.

\subsection{Adaptive Synaptic Filter}
With adaptive threshold balancing (ATB), the threshold is adjusted within the convolutional layers in order to match it to the current magnitude. However, current regulation for individual neurons will still assist the network in achieving a more stable performance.

\begin{figure}[h]
	\centering
	\includegraphics[width=0.8 \textwidth]{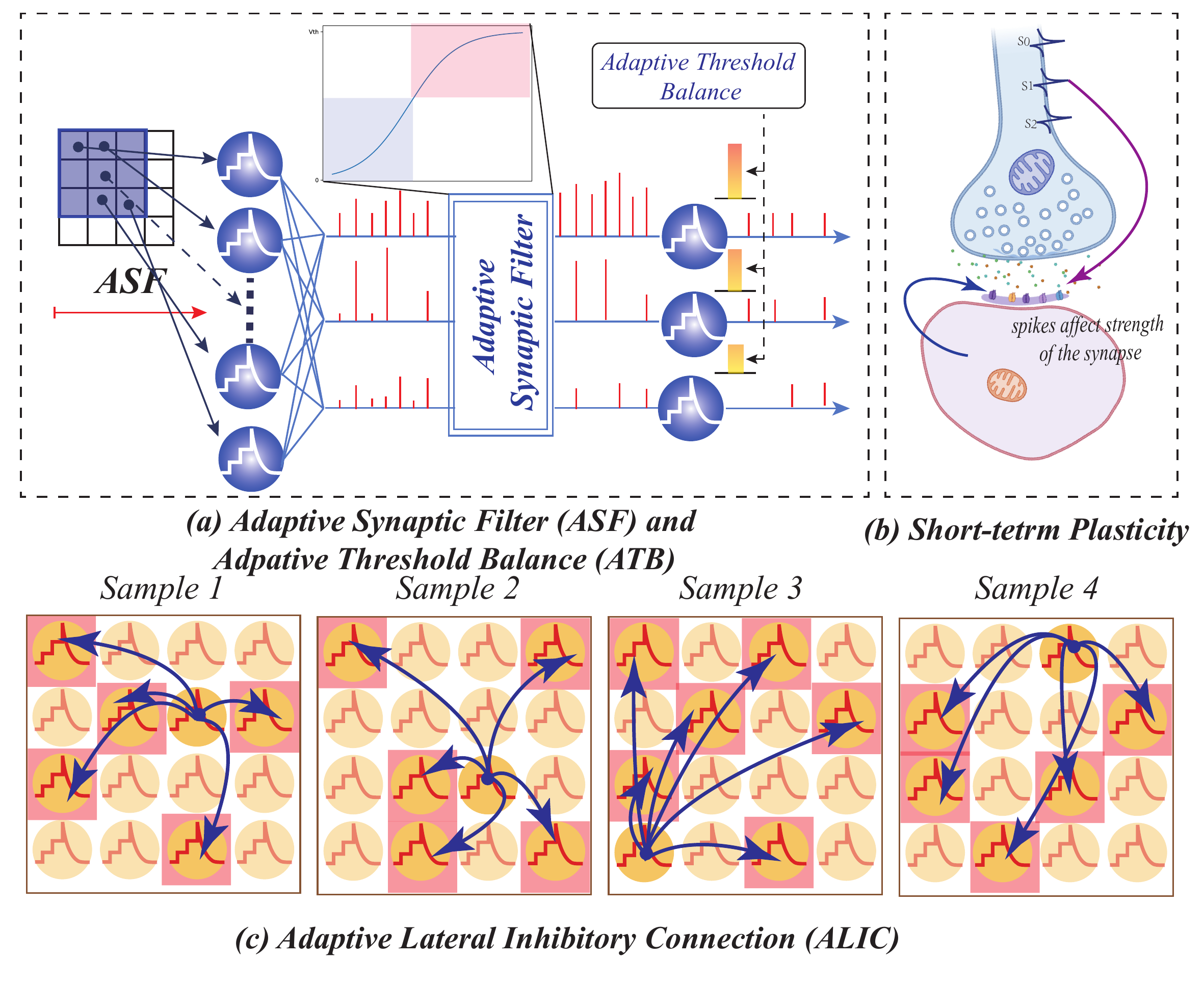} 

	\caption{\textcolor{black}{(a) The adaptive synaptic filter helps to regulate the inputs, and (b) the adaptive threshold balance helps to regulate the outputs. (c) The adaptive lateral inhibitory connection helps to suppress the same state to avoid learning repeat features.  } }
	\label{asfatbalic}
\end{figure}

The short-term plasticity (STP) of synapses affects short-term information processing at synapses, as shown in Figure \ref{asfatbalic} (b). \textcolor{black}{By increasing or decreasing the signal transmission efficiency of synapses, it can provide filter functions for the processing of information~\cite{scott2012quantifying, rotman2011short}. } Inspired by this, we construct an adaptive synaptic filter (ASF) based on the input current. Details are shown in the Equation~\ref{eqstp}.

\begin{equation}
	\label{eqstp}
	\begin{split}
		&\delta_{\mathit{asf}}(i ^{(t)}) =\frac{u^{(t)}_{\mathit{thresh}}}{1+e^{\sigma_{t}}}  \\
		&where \quad\sigma_{t}={-\alpha_{\mathit{asf}}\frac{ i ^{(t)}}{u^{(t)}_{\mathit{thresh}}}+\beta_{\mathit{asf}}}
	\end{split}
\end{equation}

\textcolor{black} { $\delta_{\mathit{asf}} $  is the function of ASF, where $u_{\mathit{thresh }}^{(t)}$ denotes the threshold at time $t$. $i  ^{(t)}$ is the input current at time $t$. $\alpha_{\mathit{asf}}$ and $\beta_{\mathit{asf}}$ are coefficients and control the function of the filter.}
The ASF adjusts the current through a nonlinear function, making the current more likely to concentrate near the threshold or resting potential. Concentrating current towards the threshold will result in more competition for the neuron. More competition will help avoid the dominance of neurons. And the current approach to resting potential will decrease the noise generated by low current strength neurons firing. 
Since ASF is calculated according to the threshold, ASF must be performed concurrently with ATB. We verify the effectiveness of ASF in section \ref{analysis}.

\subsection{STB-STDP}
\label{stbstdp}
A majority of STDP-based SNN algorithms are trained with one sample at a time, with the weights being updated at every step of the training process. However, as with the backpropagation algorithm, combining a set of samples into batches for training can reduce the convergence instability caused by deviations in the distribution of each sample. Due to the batch processing of the samples, the convergence speed of the model is also significantly increased. In addition, SNNs exhibit multiple forward propagation processes in the temporal dimension, in contrast to artificial neural networks. By aggregating multiple time steps, the network can converge more smoothly instead of updating at every step.

From this, we propose a sample-temporal batches STDP algorithm (STB-STDP). A weight update is done by integrating the information of samples and temporal information simultaneously. As the Equation \ref{eqstb} shows:

\begin{equation}
	\begin{split}
		&\Delta w_{j}=\sum \limits_{n=0}^{N_{batch}}\sum \limits_{t=0}^{T_{batch}}(x_{trace}^{n,t}-\textcolor{black}{x_{\mathit{offset}}})/(N_{batch}T_{batch})\quad if \:\: x_{\mathit{postspike}}=1\\
		&\begin{split}
			x_{trace}^{n,t}=\lambda_{+}x_{trace}^{n,t-1}+x_{\mathit{prespike}}
		\end{split}
	\end{split}
	\label{eqstb}
\end{equation}
where $N_{batch}  $is the batchsize of the input, $T_{batch}$ is the batchsize of time step.

Weights are normalized after updating, preventing them from diverging or shifting. As shown in Equation~\ref{norm_fc}. Different normalization methods are used for convolutional layers and fully connected layers. 
$A^{-}_{fc},A^{-}_{conv}$ is scale factor to control normalization. 
Additionally, a spike normalization module is added between each layer to stabilize the inputs and outputs, which limits the input range of the spike. This module captures spikes and outputs a firing frequency of up to one.  \\

\begin{equation}
	\label{norm_fc}
	fc:
	w_{j}^{(t)}=A^{-}_{fc}\frac{ w_{j}^{(t)}}{mean(w_{j}^{(t)})}
	\quad conv:
	w_{ij}^{(t)}=A^{-}_{conv}	\frac{w_{ij}^{(t)}-mean(w_{ij}^{(t)})}{std(w_{ij}^{(t)})}
\end{equation}

\section{Experiments}
\label{re}
To demonstrate the effectiveness of our model, we conduct experiments on the commonly used datasets, MNIST~\cite{lecun1998gradient} and FashionMNIST~\cite{xiao2017fashion}, CIFAR10 \cite{krizhevsky2009learning}. All the experiments are based on the structure consisting of a convolutional layer followed by a 2*2 max pooling layer, a spiking normalization layer, and a fully connected layer. Since our model is an unsupervised network, we adopt the same voting strategy as in~\cite{diehl2015unsupervised} of the output of the final layer for category prediction. And the parameters of the network are trained layer-wisely. Where $A^{-}_{fc}=0.01,A^{-}_{conv}=1$,$\alpha_{inh}=1.625$, $\theta_{init}=10$, $\alpha_{\mathit{sfa}}=0.4,  \beta_{\mathit{sfa}}=8, \alpha_{plus}=0.001, \lambda=0.99,\;\beta_{thresh}=1,\;x_{\mathit{offset}}=0.3$.
\subsection{Experimental Results}
MNIST is a digital handwriting recognition dataset widely used as a benchmark for evaluating model performance in recognition and classification tasks. The dataset contains a total of 60,000 training set samples and 10,000 test set samples. The size of each sample is 28*28 pixels. In the experiments, the examples in MNIST were normalized using direct encoding and without any form of data augmentation and preprocessing. For the MNIST dataset, we set the kernel size of the convolutional layer to 5 with 12 channels, and 6400 neurons in the fully connected layer. Timestep is 300 in our experiments. To verify the superiority of our model, we compare the results with other famous STDP-based SNN models. Un-\&Supervised denotes the former layer is trained unsupervised, while the final decision layer is trained with supervised information. As shown in Table~\ref{table1}, Our model achieves 97.9\% accuracy. Compared with \cite{diehl2015unsupervised}, which only uses the STDP, our model improves by nearly 3\%. Our model has surpassed all the unsupervised STDP-based SNNs and even some SNNs with supervised information.
\begin{table}[h]
	\caption{The performance on MNIST dataset compared with other STDP-based SNNs.}
	\centering
	\resizebox{\linewidth}{!}{
		\begin{tabular}{llrr}
			\toprule 
			Model  &Learning Method& Type&  Accuracy \\
			\midrule
			
			\cite{querlioz2013immunity} & STDP &  Unsupervised&93.5\%  \\
			\cite{diehl2015unsupervised}& STDP&  Unsupervised &95.0\%\\
			\cite{hao2020biologically}&Sym-STDP + SVM&Un-\&Supervised&96.7\%\\
			DCSNN~\cite{mozafari2019bio}&DoG + STDP + R-STDP&Un-\&Supervised&97.2\%\\
			\cite{falez2019multi}&DoG + STDP + SVM&Un-\&Supervised&98.6\%\\
			SDNN \cite{kheradpisheh2018stdp} &DoG + STDP + SVM&  Un-\&Supervised &98.4\%\\
			SpiCNN\cite{lee2018deep}&LoG+STDP&Un-\&Supervised&91.1\%\\
			\cite{tavanaei2017multi}&STDP + SVM& Un-\&Supervised&98.4\% \\
			\cite{ferre2018unsupervised}&STDP + BP&Un-\&Supervised&98.5\%\\
			SSTDP\cite{liu2021sstdp}&STDP + BP&Supervised&98.1\%\\
			VPSNN~\cite{zhang2018plasticity}&Equilibrium Propagation + STDP&Supervised&98.5\%\\
			CBSNN~\cite{shi2020curiosity}&VPSNN + Curiosity&Supervised&98.6\%\\
			BP-STDP~\cite{tavanaei2019bp}&STDP-Based BP&Supervised&97.2\%\\
			GLSNN~\cite{zhao2020glsnn}&Global Feedback + STDP&Supervised&98.6\%\\
			\midrule
			Ours&ASF + ATB + ALIC + STB-STDP&Unsupervised&97.9\%\\
			\bottomrule
		\end{tabular}
	}
	\label{table1}
\end{table}

FashionMNIST is more complex than MNIST within cloths and shoes as the samples. Both the shape and the size of the data are the same as those obtained by MNIST. The kernel size for the convolutional layer is set at 3 with 64 channels, and 6400 neurons in the fully connected layer. The timestep is the same as that used in MNIST. As seen in Table~\ref{table2}, our model succeeded in achieving 87.0\% accuracy and outperformed most STDP-based SNNs. Although the performance of GLSNN is higher than ours, it introduces global supervised connections, while our network has no supervision information. 

\begin{table}[h]
	\caption{The performance on FashionMNIST dataset compared with other STDP-based SNNs.}
	\centering
	\resizebox{1.0\linewidth}{!}{
		\begin{tabular}{llrr}
			\toprule 
			Model  &Learning Method& Type&  Accuracy \\
			\midrule
			FSpiNN \cite{putra2020fspinn}&STDP&Unsupervised&68.8\%\\
			\cite{rastogi2021self}&A-STDP&Unsupervised&75.9\%\\
			\cite{hao2020biologically}&Sym-STDP&Supervised&85.3\% \\
			VPSNN~\cite{zhang2018plasticity}&Equilibrium Propagation + STDP&Supervised&83.0\%\\
			CBSNN~\cite{shi2020curiosity}&VPSNN + Curiosity&Supervised&85.7\%\\
			GLSNN~\cite{zhao2020glsnn}&Global Feedback + STDP&Supervised&89.1\%\\
			\midrule
			Ours&ASF + ATB + ALIC + STB-STDP&Unsupervised&87.0\%\\
			\bottomrule
		\end{tabular}
	}
	\label{table2}
\end{table}

\subsection{Result on CIFAR10}

\begin{figure} 
	\centering
	\includegraphics[width=0.49\textwidth]{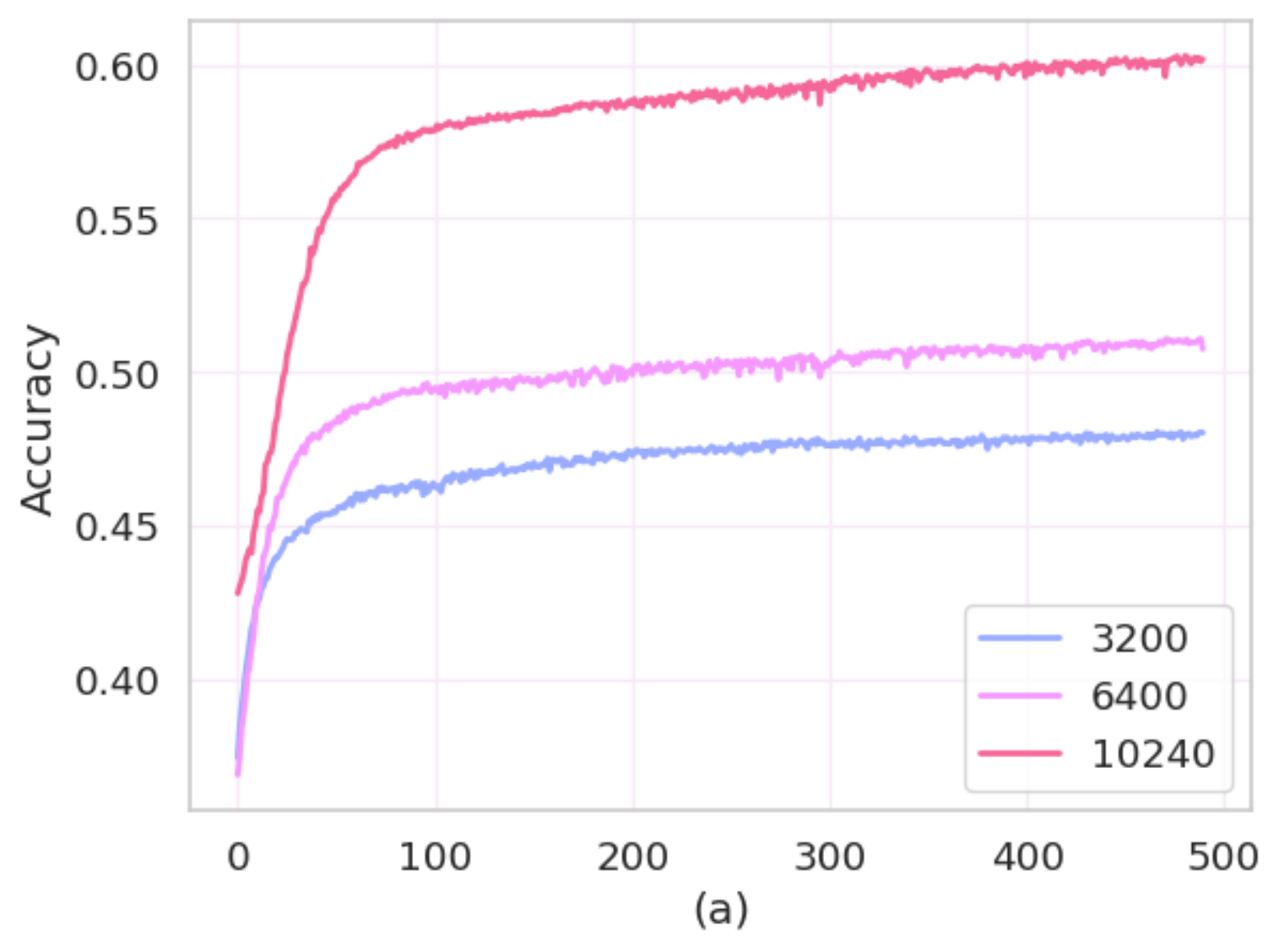}
	\includegraphics[width=0.49\textwidth]{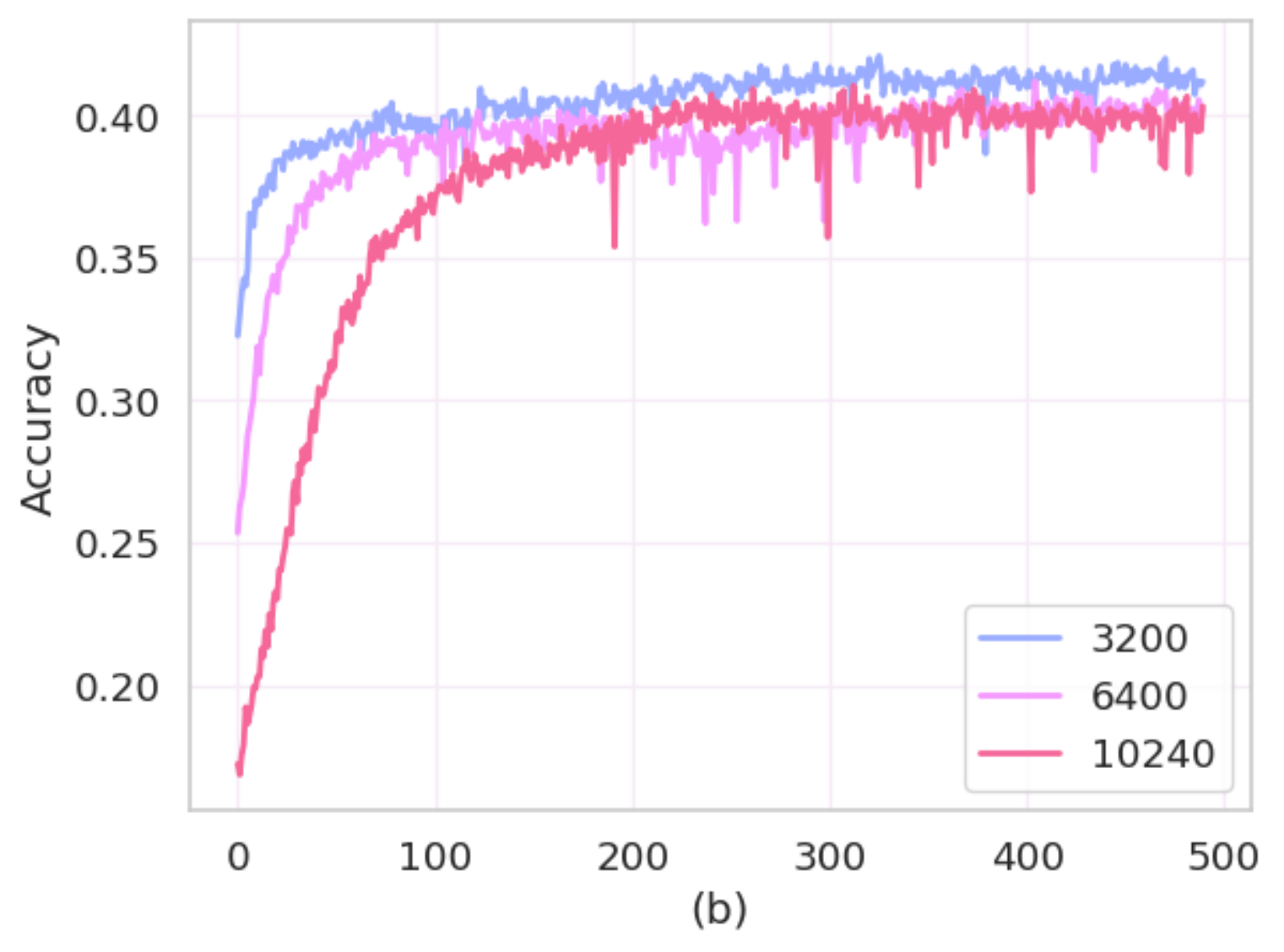}
	\caption{\textcolor{black}{ The accuracy curve of the model on cifar10. We conducted experiments with 3200, 6400, and 10240 neurons in the fully connected layer. We visualize (a) the accuracy curve for the training set and (b) the accuracy curve for the test set.}}
	\label{curve}
\end{figure}

Only a few STDP-based SNN algorithms are capable of classifying the FashionMNIST dataset and achieving impressive results. As far as we know, few unsupervised STDP-based models run experiments on more complex datasets, such as CIFAR10. Trying to maintain high performance on more complex datasets will be our exploration direction. First, we used RGB data and the image size is 32x32. We use a 5x5 convolution kernel with 3 input channels and 64 output channels in the convolutional layer. Meanwhile, 3200 neurons are used in the fully connected layer. After training, our model achieved 42.36\% accuracy on the CIFAR10 dataset. We visualize the accuracy curves of our model with 3200, 6400, and 10240 neurons in the fully connected layer. The highest accuracy was obtained for the test set at the number of neurons of 3200.

\subsection{Discussion}
\begin{figure}[!h]
	\centering
	\includegraphics[width=1.0\textwidth]{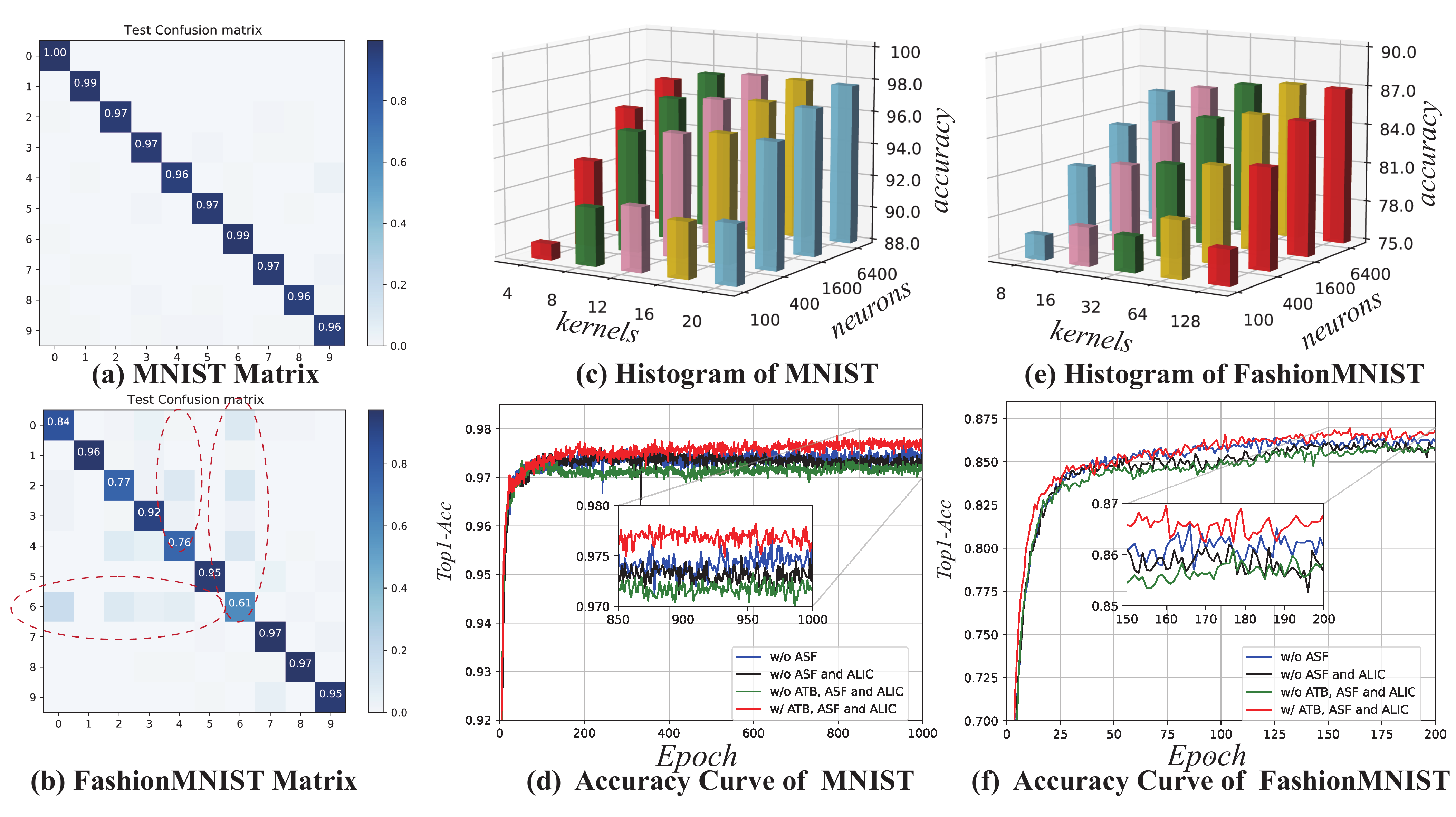}
	\caption{\textcolor{black}{(a,b) The confusion matrix of our model in MNIST and FashionMNIST. (c,e) The histograms of the performance of our model on MNIST and FahsionMNIST with the different numbers of convolutional kernels and voting neurons. (d,f) The test accuracy curves of our model with and without the modules we design.}}
	\label{accfull}
\end{figure}

In order to determine what causes misclassified samples, we visualized the classification confusion matrix of our model on the MNIST and FashionMNIST datasets, respectively. As shown in the Figure \ref{accfull}. There is not much distinction between the categories on MNIST's matrix. However, FashionMNIST's matrix showed lower accuracy on four categories of 0,2,4,6, corresponding to 't-shirt', 'pullover', 'coat', and 'shirt', respectively. There is a high degree of similarity between images in these four categories, which makes it more difficult for the model to classify them accurately.

We explore the impact of different parameter settings on network performance. In convolutional layers, we changed the number of kernels, while in fully connected layers, we changed the number of neurons. As shown in Figure ~\ref{accfull}. The different number of channels and the number of neurons in the fully connected layer are combined, respectively. The height of the histogram represents the final accuracy of the model. There is a direct correlation between performance and the number of voting neurons set. Because more voting neurons will provide more predictions, combining all predictions, the network will produce more accurate results. However, this correlation does not exist for the setting of the number of convolution kernels. The best performance for MNIST is achieved when the number of kernels is set to 12.

In contrast, FashionMNIST requires more convolution kernels due to its complexity, and it works best when the number of kernels is set to 64. More convolution kernels cannot extract more features, because repeated features will appear in large numbers. In addition, the fully connected layer also needs more connections due to the increased number of kernels because more connections increase the complexity of the network and the difficulty of convergence.

To fully illustrate the impact of each module of our model on the results,  we conduct ablation experiments on two datasets separately. \textcolor{black}{We conducted experiments with the settings "w/o ASF", "w/o ASF and ALIC", and "w/o ASF, ALIC, and ATB", respectively.  "w/o" denotes that we remove the relevant module.} We  show the convergence curve of the model on the test set, as shown in Figure \ref{accfull}. All experiments included STB-STDP to ensure the efficiency of the experiment.  It would take a much longer time for the model to run without it. Adding each module can assist in improving the model's learning ability and its ability to achieve higher accuracy on the MNIST and FashionMNIST datasets.

To better illustrate the role of lateral inhibition, we added "softmax" for comparison. "softmax" is a commonly used function that can serve as a lateral inhibition. For the experiments, we removed the lateral inhibition module and performed the experiments on the MNIST dataset with the same model structure. The convolutional layer contains 12 5x5 convolutional kernels and 6400 neurons in the fully connected layer. We have trained the model with the same experimental setup. The model reached 28.26\%. Then we added the softmax layer after the input currents in the convolutional layer and the fully connected layer. Also, to ensure that the currents are of the same scale, we multiplied by the maximum value of the current. After training, the model obtained 78.16\%. This shows that the softmax has a certain effect. However, lateral inhibition can help the model to achieve the inhibition in a better way.

\label{analysis}

\subsection{Experiment for Small Samples}
\begin{table}[h]
	\caption{The performance of our model compared with ANN on MNIST dataset with different training samples.}
	\centering
	\resizebox{0.5\linewidth}{!}{
		\begin{tabular}{lrrrr}
			\toprule 
			samples&200  &100&50&10 \\
			\midrule	
			ANN&79.77\%&71.40\%&68.72\%&47.12\%\\
			Ours&81.45\%&75.44\%&72.88\%&51.45\%\\
			\midrule
			&1.68\%&4.04\%&4.16\%&4.33\%\\
			\bottomrule
		\end{tabular}
	}
	\label{table3}
	
\end{table}

\textcolor{black}{In contrast to the backpropagation algorithm, our network trained with the STDP unsupervised algorithm is more adaptable to different number of samples, especially in small-sample task, and shows superior performance compared to supervised algorithms with the same structure.} There are only a very small number of samples in the whole train set in small-samples task, which differs from few-shot learning. To this end, to illustrate the ability of our model to train on small samples tasks, a very small number of samples are randomly selected from the MNIST dataset. The number of samples per class ranges from 20 to 10 to 5, in the most extreme case, to 1. At the same time, we also designed an artificial neural network model with the same structure for comparison. The model consists of 5x5 convolutions with 12 channels, a maxpool layer, a fully connected layer of 6400 neurons, and a final fully connected classifier. The backpropagation algorithm is used to train this ANN model.

As shown in the Table~\ref{table3}, our model has better performance than the ANN model. As the number of training samples gradually decreases, the performance gap between our model and ANN gradually increases.  When there is only 1 training sample per class, our model outperforms ANN by 4.43\%. It fully illustrates that our model requires only a small number of samples to achieve high performance compared to artificial neural networks that require a large amount of labeled training data.

\subsection{Visualization}
\begin{figure}[!]
	\centering
	
	\includegraphics[width=0.95\textwidth]{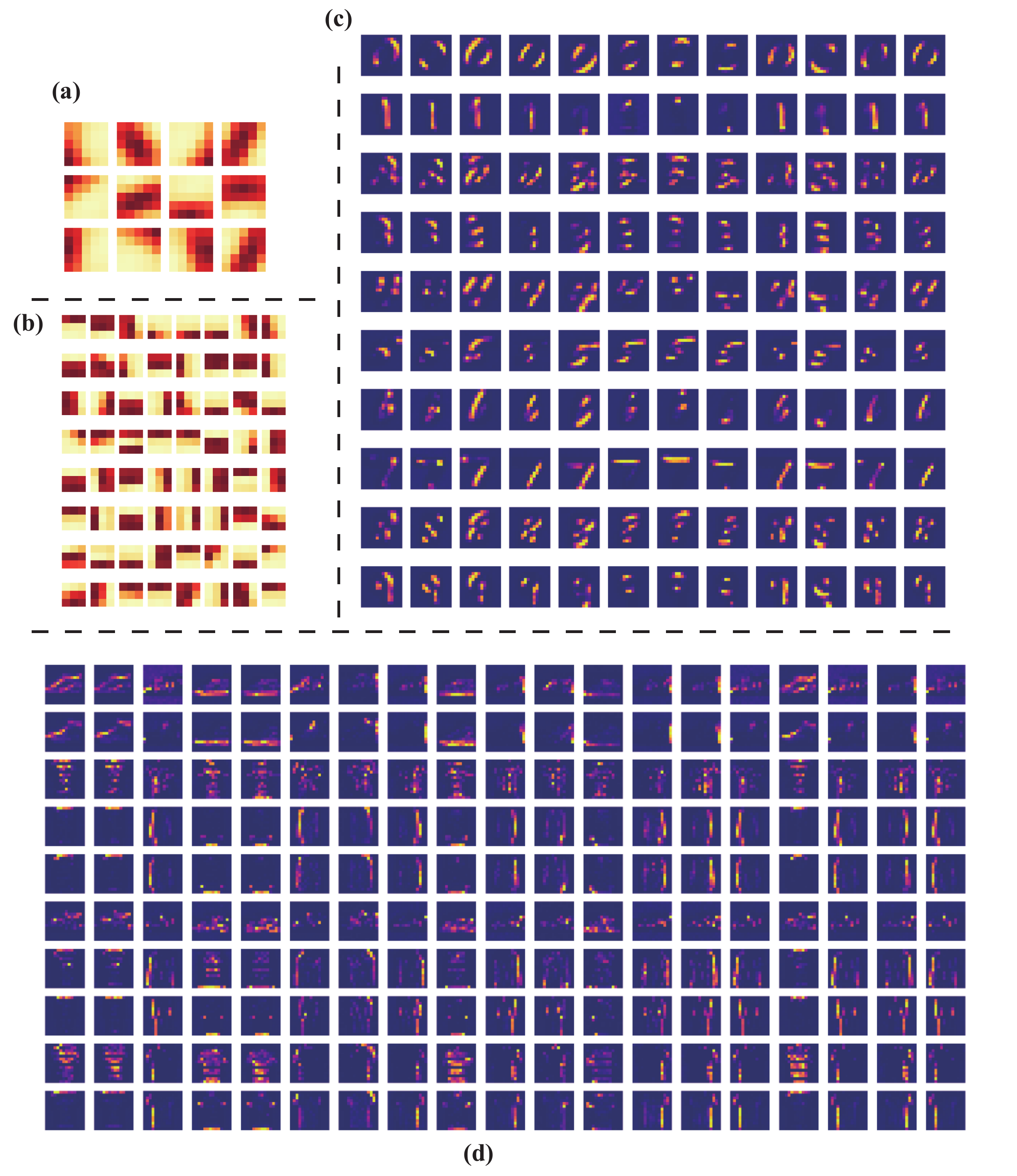} 
	\caption{ \textcolor{black}{Visualization of the weights of the model. (a) represents the convolutional layers trained on MNIST. It contains 12 channels, and we visualize the weights of each kernel. (b) Convolutional layer trained on FashionMNIST. It contains 64 kernels. (c) Fully connected layer on MNIST. We randomly selected 10 neurons from the corresponding categories. The corresponding weights are visualized. Each of these rows shows one category, (d) Fully connected layer on the FashionMNIST dataset. We visualized a portion of these weights, where each row represents the weight of neurons corresponding to a category.}}
	\label{vis}
\end{figure}
To illustrate the feature extraction capability of our model, we visualize the weights of different layers. Figure \ref{vis} a and b shows the weight of the convolutional layer on the MNIST and FashionMNIST dataset respectively. The convolutional kernels capture simple features such as edges, lines. With the introduction of our adaptive lateral inhibitory connections, our network does not have a large number of repeated features. Figure~\ref{vis} c and d show the weight of the fully connected layer. According to the label assigned to the neuron, we visualize the weight of ten categories, and each row represents a category. It can be seen that the fully connected layer automatically combines the features of the convolutional layers to form higher-level semantic representations. For MNIST, with the combination of simple features, the different numbers can be easily classified. While FashionMNIST is more complex, it is not easy to distinguish similar objects such as Shirts and T-Shirt. In future work, we will consider introducing more biologically plausible rules to improve the performance of our model.

\section{Conclusion}
Spiking neural networks (SNNs) trained with STDP alone are inefficient and hardly achieve a high performance of SNN. In this paper, we design an adaptive synaptic filter and introduce the adaptive threshold balance to enrich the representation ability of SNNs. We also introduce an adaptive lateral inhibitory connection to help the network learn richer features. We design a samples temporal batch STDP (STB-STDP), which updates weights based on multiple samples and moments. By integrating the above three adaptive mechanisms and STB-STDP,  we have achieved current state-of-the-art performance for unsupervised STDP-trained SNNs on MNIST and FashionMNIST. Further, we tested on the more complex CIFAR10 dataset, and the results fully illustrate the superiority of our algorithm. Since we consider more of the unsupervised learning rules, it does not obtain significant improvement when it is extended to deep networks. In future work, we intend to consider more learning methods of the human brain, such as dopamine-based regulation of the reward mechanism. Also, We will introduce more brain-like structures, such as global feedback connections.

\label{con}

\section*{Acknowledgement}
This study was supported by National Key Research and Development Program (Grant No.2020AAA0107800), the Strategic Priority Research Program of the Chinese Academy of Sciences (Grant No.XDB32070100).

\bibliography{nn_2022}

\end{document}